\documentclass{article}

\usepackage{microtype}
\usepackage{graphicx}
\usepackage{subfigure}
\usepackage{booktabs} 
\usepackage{microtype}
\usepackage{footnote}
\usepackage{amsmath}

\usepackage{hyperref}

\newcommand{\rlstm}{{\it r}-LSTM}
\newcommand{\plstm}{{\it p}-LSTM}
\usepackage[accepted]{icml2018}

\icmltitlerunning{Learning Longer-term Dependencies in RNNs with Auxiliary Losses}

\begin{document}

\twocolumn[
\icmltitle{Learning Longer-term Dependencies in RNNs with Auxiliary Losses}



\begin{icmlauthorlist}
\icmlauthor{Trieu H. Trinh\footnotemark}{}
\icmlauthor{~Andrew M. Dai}{} 
\icmlauthor{~Minh-Thang Luong}{} 
\icmlauthor{~Quoc V. Le}{} 
\\
  {\tt \{thtrieu, adai, thangluong, qvl\}@google.com}
\end{icmlauthorlist}

\icmlkeywords{Machine Learning, ICML}

\vskip 0.3in
]

\footnotetext{Work done as a member of the Google Brain Residency program (\url{g.co/brainresidency}.)

\textit{Proceedings of the
$\mathit{35}^{th}$ International Conference on Machine Learning},
Stockholm, Sweden, PMLR 80, 2018.
Copyright 2018 by the author(s).
}

\begin{abstract}
Despite recent advances in training recurrent neural networks (RNNs), capturing long-term dependencies in sequences remains a fundamental challenge.
Most approaches use backpropagation through time (BPTT), which is difficult to scale to very long sequences. 
This paper proposes a simple method that improves the ability to capture long term dependencies in RNNs by adding an unsupervised auxiliary loss to the original objective. 
This auxiliary loss forces RNNs to either reconstruct previous events or predict next events in a sequence, making truncated backpropagation feasible for long sequences and also improving full BPTT. 
We evaluate our method on a variety of settings, including pixel-by-pixel image classification with sequence lengths up to 16\,000, and a real document classification benchmark. 
Our results highlight good performance and resource efficiency of this approach over competitive baselines, including other recurrent models and a comparable sized Transformer. 
Further analyses reveal beneficial effects of the auxiliary loss on optimization and regularization, as well as extreme cases where there is little to no backpropagation. 
\end{abstract}

\section{Introduction}

\begin{figure}[t!]
\centering
\includegraphics[width=\columnwidth]{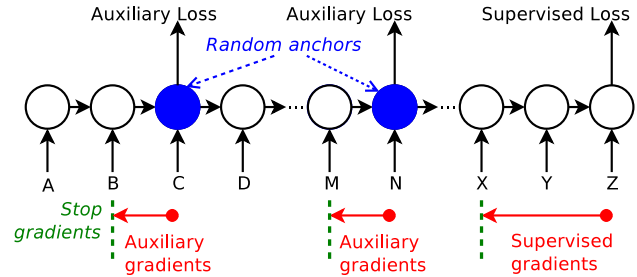}
\caption{An overview of our method. The auxiliary loss improves the memory of the recurrent network such that the number of steps needed for the main task's BPTT is small.}
\label{fig:overview}
\end{figure}
Many important applications in artificial intelligence require the understanding of long term dependencies between events in a sequence. For example, in natural language processing, it is sometimes necessary to understand relationships between distant events described in a book to answer questions about it. Typically, this is achieved by gradient descent and BPTT~\cite{Rumelhart86} with recurrent networks. Learning long term dependencies with gradient descent, however, is difficult because the gradients computed by BPTT tend to vanish or explode during training~\cite{hochreiter2001gradient}. Additionally, for BPTT to work, one needs to store the intermediate hidden states in the sequence. The memory requirement is therefore proportional to the sequence length, making it difficult to scale to large problems.

Several promising approaches have been proposed to alleviate the aforementioned problems. First, instead of using the vanilla recurrent network, one can use Long Short-Term Memory (LSTM)~\cite{hochreiter1997long}, which is designed to improve gradient flow in recurrent networks. In addition, one can also use gradient clipping~\cite{pascanu2013difficulty} to stabilize the training of the LSTM. Finally, to reduce the memory requirement, one can either store the hidden states only periodically~\cite{gruslys2016memory, chen2016training}, use truncated BPTT, or use synthetic gradients~\cite{Jaderberg16synthGrad}.

\begin{figure*}[tb]
\centering
\includegraphics[width=0.8\textwidth]{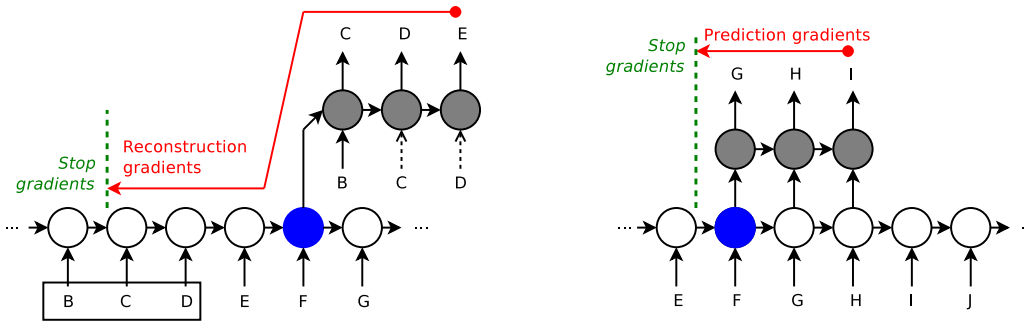}
\caption{An overview of our methods. For each random anchor point, say F, we build an auxiliary loss at its position. \textbf{Left}: We predict a random subsequence BCD that occurs before F. B is inserted into a decoder network to start the reconstruction, while C and D is optionally fed. \textbf{Right}: We predict the subsequence GHI by stacking an auxiliary RNNs on top of the main one. Gradients from auxiliary loss is truncated in both cases to keep the overall cost of BPTT constant.}
\label{fig:method}
\end{figure*}

Convolutional neural networks also mitigate the problem of long-term dependencies since large kernel sizes and deep networks such as ResNets~\cite{he2016deep} allow long-term dependencies to be learnt across distant parts of an image. However, this is a fundamentally different kind of architecture that has other tradeoffs. For example, the entire input  (an image or sequence) and the intermediate activations of the model must be stored in memory during training. At inference time, typical CNNs also need $O(n)$ storage where $n$ is the size of the input.\footnote{A convolutional layer is often followed by a reduction layer to reduce the input size by a constant factor.} The Transformer~\cite{vaswani2017attention} has a similar issue, though somewhat magnified since computation for training and inference requires random access to storage that is $O(n)$.

RNNs therefore have the advantage where assuming a fixed BPTT length of $l$, training requires $O(l)$ storage. This is commonly the case when training language models on the PTB dataset~\cite{penntreebank}, where the state is never reset over the entire 1 million token sequence. Therefore, in theory the RNN can learn relationships across this extremely long distance. Furthermore, inference in RNNs also requires $O(1)$ storage since RNNs do not need to `look back'.

In this paper, we propose an orthogonal technique to further address the weakness of recurrent networks purely relying on BPTT. Our technique introduces an unsupervised auxiliary loss to the main supervised loss that reconstructs/predicts a random segment in the sequence before/after an anchor point. This enables learning with only need a few BPTT steps from the supervised loss.

Our results show that unsupervised auxiliary losses significantly improve optimization and generalization of LSTMs. Moreover, using this technique, one does not have to perform lengthy BPTT during training to obtain good results. Our method, therefore, lends itself to very long sequences where vanishing/exploding gradients as well as the cost of lengthy BPTT become critical bottlenecks. 

In our experiments where sequences of up to 16\,000 elements is processed, LSTMs with auxiliary losses can train much faster and with less memory usage, while training LSTMs with full backprop becomes very difficult.

\section{Related works}

As learning long term dependencies with recurrent networks is an important problem in machine learning, many approaches have been proposed to tackle this challenge. Well known approaches include recurrent networks with special structures~\cite{el1996hierarchical,sperduti1997supervised,frasconi1998general,socher11,chan2016listen}, Long Short-Term Memory Networks~\cite{hochreiter1997long,gers1999learning,graves2013generating}, Gated Recurrent Unit Networks~\cite{cho2014learning,chung2014empirical}, multiplicative units~\cite{wu2016multiplicative}, specialized optimizers~\cite{martens2011learning,kingma2014adam}, identity initialization and connections~\cite{mikolov2014learning,le2015simple,he2016deep}, highway connections~\cite{zilly2016recurrent}, orthogonal- or unitary-constrained weights~\cite{white2004short,henaff2016recurrent,arjovsky2016unitary}, dilated convolutions~\citep{salimans2017pixelcnn++}, connections~\cite{koutnik2014clockwork} and attention mechanisms~\citep{bahdanau2014neural,luong2015effective,vaswani2017attention}. A more recent approach is to skip input information at certain steps~\cite{yu2017learning,seo2017neural,campos2017skip}. As training very long recurrent networks is memory-demanding, many techniques have also been proposed to tackle this problem~\cite{chen2016training,gruslys2016memory,Jaderberg16synthGrad}. We propose methods that are orthogonal to these approaches, and can be used in combination with them to improve RNNs. 

Our work is inspired by recent approaches in pretraining recurrent networks~\cite{dai2015semi,ramachandran2016unsupervised} with sequence autoencoders or language models. Their work, however, focuses on short sequences, and using pretraining to improve generalization of these short recurrent networks. In contrast, our work focuses on longer sequences, and studies the effects of auxiliary losses in learning long term dependencies.

Combining auxiliary losses and truncated BPTT is also described in the context of online learning~\cite{schmidhuber1992learning}, where the main network learns to predict the concatenation of its next input token, the target vector, and distilled knowledge from an auxiliary network. The auxiliary network only predicts the sequence of tokens that is not predicted correctly by the main network. This shorter sequence is termed the {\it compressed history} and is argued to be suffice for good classification. In variational inference setting, \cite{goyal2017z} also propose reconstruction of the states from a backward running recurrent network as an auxiliary cost to help with training on long sequences.

\section{Methodology}

An overview of our methods is shown in Figure~\ref{fig:overview}. Let us suppose that the goal is to use a recurrent network to read a sequence and classify it. We propose to randomly sample one or multiple anchor positions, and insert an unsupervised auxiliary loss at each of them.

\subsection{Reconstruction auxiliary loss}

In reconstructing past events, we sample a subsequence before the anchor point, and insert the first token of the subsequence into a decoder network; we then ask the decoder network to predict the rest of the subsequence. The whole process is illustrated in Figure~\ref{fig:method}-left.

Our intuition is that if the events to be predicted are close enough to the anchor point, the number of BPTT steps needed for the decoder to reconstruct past events can be quite small. Furthermore, with this training, the anchor points serve as a temporary memory for the recurrent network to remember past events in the sequence. If we choose enough anchor points, the memory is built over the sequence such that when we reach sequence end, the classifier remembers enough about the sequence and can do a good job at classifying it. Consequently, the classifier only needs a few backpropagation steps to fine-tune the LSTM's weights, since good embeddings of the input sequence has been learnt by optimizing the auxiliary objective.

\subsection{Prediction auxiliary loss}

Another auxiliary loss of consideration is analogous to Language Modelling loss, illustrated in Figure~\ref{fig:method}-right. In this case, we ask the decoder network to predict the next token given the current one sequentially, over a subsequence starting from the anchor point. This type of unsupervised auxiliary loss is first examined by ~\citet{dai2015semi}, where it is applied over the whole input sequence. In our experiments, however, we are interested in scalable schemes of learning long term dependencies, we therefore only apply this loss on a subsequence after the random anchor point.

\subsection{Training}

We name the former method \rlstm{}, the later \plstm{} (which respectively stand for {\it reconstruct-} and {\it predict-}LSTM) and train them in two phases. The first is pure unsupervised pretraining where only the auxiliary loss is minimized. In the second phase, semi-supervised learning is performed where we minimize the sum of the main objective loss $L_{\textrm{supervised}}$ and our auxiliary loss $L_{\textrm{auxiliary}}$. The auxiliary LSTM that performs reconstruction is trained with Scheduled Sampling~\cite{bengio2015scheduled}.

\subsection{Sampling frequency and subsequence length}

By introducing the auxiliary losses over subsequences of the input, one introduces extra hyper-parameters. The first indicates how frequently one should sample the reconstruction segments, the others indicate how long each segment should be. Denoting the former \(n\), and the later \(\{l_i\}_{i=1}^n\), we obtain the auxiliary loss as follows:$
    L_{\textrm{auxiliary}} = \frac{\sum_{i=1}^nL_i}{\sum_{i=1}^nl_i}$

Where $L_i$ denotes the loss evaluated on the $i^{th}$ sampled segment, and is calculated by summing losses on all predicted tokens ($\textrm{TokenLoss}_t$) in that segment:
    $L_{i} = \sum_{t=1}^{l_i}\textrm{TokenLoss}_t$
For sequences of characters, each $\textrm{TokenLoss}_t$ is the cross-entropy loss between the ground truth one-hot vector and the prediction produced by our decoder network. For other types of input, we treat each token as a continuous, multi-dimensional real vector and perform $L_2$ distance minimization.

Tuning hyper-parameters is known to be very expensive, especially so when training RNNs on very long sequences. We therefore set all sampled segments to the same length: $l_i = l \ \forall i$, and sample at frequency $n=1$ in most experiments. Tuning these hyper-parameters is also explored in cases where sequence length is relatively short. In later experiments, we show that the tuned values generalize well to much longer input sequences.

\section{Experiments} 

To evaluate the effectiveness of our models, we consider a wide variety of datasets with sequences of varying lengths from 784 to 16384. Our first benchmark is a {\it pixel-by-pixel image classification} task on MNIST in which pixels of each image are fed into a recurrent model sequentially before a prediction is made. This dataset was proposed by \citet{le2015simple} and has now become the most popular benchmark for testing long term dependency learning.\footnote{No symbol was added to indicate the end of each row of pixels.}

Beside MNIST, we also explore pMNIST, a harder version, where each pixel sequence is permuted in the same way. Permuting pixels breaks apart all local structures and creates even more complex dependencies across various time scales. To test our methods on a larger dataset, we include pixel-by-pixel CIFAR10 (no permutation). Additionally, to perform control experiments with several scales of sequence lengths, we use the StanfordDogs dataset \citep{stanforddogs} which contains large images categorized to 120 dog breeds. All images are scaled down to 8 different sizes from 40$\times$40 to 128$\times$128 before being flattened into sequences of pixels without permutation. This setup results in sequences of lengths up to 16\,000, which is over 20 times longer than any previously used benchmark of this flavor.



\begin{table}[]
\caption{Datasets and average sequence length.}
\label{image_statistics}
\vskip 0.15in
\begin{center}
\begin{small}
\small
\begin{tabular}{lccc}
\toprule
Dataset & Mean length & \# classes & Train set size\\
\midrule
MNIST     & 784 &  10 & 60K\\
pMNIST  & 784 & 10 & 60K\\
CIFAR10 & 1024 & 10 & 50K\\
StanfordDogs\footnotemark  & 1600 -- 16384 & 120 & 150K\\
DBpedia & 300 & 14 & 560K\\
\bottomrule
\end{tabular}

\end{small}
\end{center}
\vskip -0.1in
\end{table}
\footnotetext{We follow the procedure suggested in \citet{SermanetFR14} to obtain a larger training set, while keeping the same test set. All images are scaled down to 8 different sizes from 40$\times$40 to 128$\times$128 before being flattened into sequences of pixels.}


Lastly, we explore how well truncated BPTT and the auxiliary losses can perform on a real language task, where previous RNN variants have already reported remarkable accuracy. For this task, the DBpedia character level classification task is chosen as it is a large-scale dataset (with 560K training examples) and has been well benchmarked by~\citet{dai2015semi}.
We follow the procedure suggested in ~\citet{zhang2015text} to normalize the dataset.

A summary of all datasets being used is presented in Table~\ref{image_statistics}.

\subsection{Model Setup}
We use a single-layer LSTM with 128 cells and an embedding size of 128 to read the input sequence. 
For the supervised loss, the final state of the main LSTM is passed through a two-layer feedforward network (FFN) with 256 hidden units, before making a prediction. We apply drop-connect~\cite{wan2013regularization} with probability 0.5 on the second layer. For the auxiliary losses, we use a two-layer LSTM in which the bottom layer is initialized from the current state of the main classification LSTM, while the top one starts with zero state. When reconstructing image pixels, a two-layer FFN (256 units, drop-connect 0.5 on second layer) is applied on top of the auxiliary LSTM per timestep.

Our RNNs are trained using the RMSProp optimizer~\cite{Tieleman2012} with batch size of 128. Unsupervised pretraining is done in 100 epochs with initial learning rate of 0.001, which is halved to 0.0005 halfway through pretraining. For the semi-supervised phase, the same learning rate is halved every 300 epochs until training reaches 1000 epochs. Scheduled sampling for auxiliary LSTMs is annealed linearly to zero after 100\,000 training steps.

As we scale our methods to various input lengths, we make sure that backpropagation cost is constant regardless of the input length. Specifically, gradients are truncated to 300 time steps for both the supervised and auxiliary losses.\footnote{All models are implemented in TensorFlow~\cite{tensorflow2015-whitepaper}. Truncated gradients are achieved using the built-in \texttt{tf.stop\_gradient} op.}
For the auxiliary losses, we choose the simplest setup of sampling $n{=}1$ segment of length $l{=}600$ per training example. In Section~\ref{sec:sample_frequency}, we will explore different values for $n$ and $l$.

As a complement to results from purely recurrent models, in Section~\ref{sec:transformer}, we will also compare our models with Transformer \cite{vaswani2017attention}. Transformer is a completely different paradigm of processing sequences that sidesteps the difficulties of BPTT through the use of self-attention. Such advantage is achieved at the cost of $O(n)$ working memory during both training and inference compared to $O(1)$ for RNNs. Even though our main interest is to improve over recurrent models, we include these results to study how scalable the self-attention mechanism is. 

We use Tensor2Tensor\footnote{\url{https://github.com/tensorflow/tensor2tensor}} to train Transformer models with an off-the-shelf configuration that has a comparable number of parameters as our RNNs (0.5M weights)\footnote{\texttt{transformer\_tiny}.}. A simple setting for classification is adopted where the Transformer output vectors is average-pooled and fed into a two-layer FFN before making predictions, as done in our RNNs.


\subsection{Main results}

\subsubsection{MNIST, pMNIST, and CIFAR10}
We first explore sequences of length no longer than 1000 on MNIST, pMNIST and CIFAR10. Besides results from previous works on pixel MNIST and permuted MNIST (pMNIST) such as \citet{le2015simple,arjovsky2016unitary}, we evaluated a fully trained LSTM and an LSTM trained with only 300 steps of BPTT as the main baselines to see how much disadvantage truncating classification gradients might cause. At this stage, it is also affordable to include test accuracies from both truncated and fully-trained \rlstm{} and \plstm{} for a more complete result.

\begin{table}[htb]
\caption{Test accuracy (\%) on MNIST, pMNIST, and CIFAR10.}
\label{mnist_cifar}
\vskip 0.15in
\begin{center}
%
\resizebox{8cm}{!}{
\begin{tabular}{lccc}
\toprule
 & MNIST & pMNIST & CIFAR10 \\
\midrule
iRNN \cite{le2015simple} & 97.0 & 82.0 & N/A \\
uRNN \cite{arjovsky2016unitary} & 95.1 & 91.4 & N/A \\
LSTM Full BP & 98.3 & 89.4 & 58.8 \\
LSTM Truncate300 & 11.3 & 88.8 & 49.0 \\
\midrule
\rlstm{} Truncate300 & 96.4 & 92.8 & 65.9 \\
\plstm{} Truncate300 & 95.4 & 92.5 & 64.7 \\
\textbf{\rlstm{} Full BP} & \textbf{98.4} & \textbf{95.2} & \textbf{72.2} \\
\plstm{} Full BP & 98.0 & 92.8 & 67.6 \\
\bottomrule

\end{tabular}
}
%
\end{center}
\vskip -0.1in
\end{table}

An overview of Table~\ref{mnist_cifar} shows that our proposed auxiliary losses produce gradually larger improvements moving from MNIST to pMNIST and CIFAR10. On pixel-by-pixel MNIST, our truncated LSTM baseline is nearly untrainable, with only 11.3\% accuracy. This is due to the fact that gradients back-propagated from the loss can only reach largely non-informative solid pixels near the end of the sequence. Despite this detrimental effect of gradient truncation, the proposed unsupervised losses bring \rlstm{} and \plstm{} on par with fully trained RNNs like uRNN and LSTM.

On permuted pMNIST where more complex long-range dependencies is put to the test, \rlstm{} and \plstm{} easily outperform the fully trained LSTM baselines as well as a fully trained uRNN, while using less than half the number of gradients from the classification loss.

On CIFAR10, we observe an even greater discrepancy, where \rlstm{} is followed closely by \plstm{} in accuracy, while a fully trained LSTM is more than 7\% lower in absolute accuracy. 

With fully backpropagated gradients from classification loss, we obtain the best accuracy across all datasets against other recurrent models. Notably on the two harder benchmarks pMNSIT and CIFAR10, \rlstm{} outperforms a fully-trained LSTM by a large margin.

\subsubsection{StanfordDogs}

\begin{figure}[tb]
\includegraphics[width=1\columnwidth, clip=true, trim= 10 40 60 80]{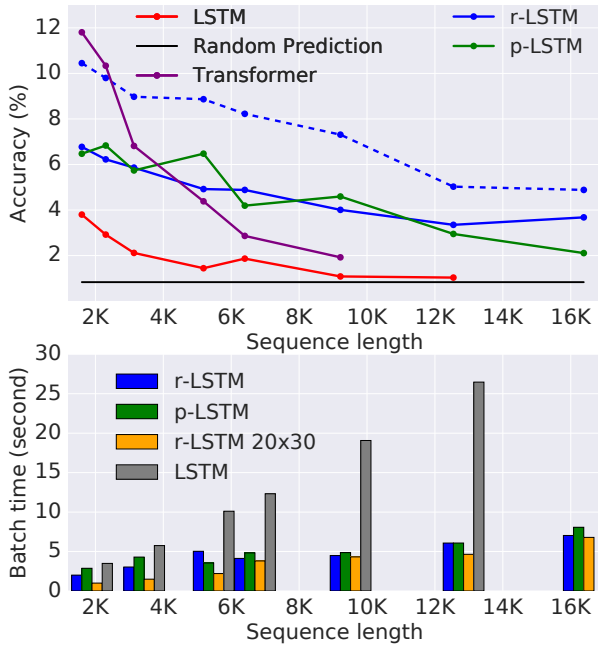}
\caption{Top: Test accuracy on StanfordDogs resized to 8 levels of sequence length. Bottom: Time to run a single mini batch of 128 training examples, measured in second.}
\label{fig:growing_lengths}
\end{figure}

So far, our experiments give hints that \rlstm{} and \plstm{} scale better in performance when input sequences get longer and more complex. Next, we present how this trend elaborates when input sequences extend up to an order of magniwtude higher -- over 10\,000 steps. As presented earlier, we use the dataset StanfordDogs resized down to 8 levels of sequence lengths and test the models on all levels.

As this range, training is expensive in terms of time and computational resources, especially so with LSTMs where parallelization over the time dimension is not possible. We therefore restrict each training session to the same amount of resource (a single Tesla P100 GPU) and report infeasible whenever a mini-batch of one training example can no longer fit into memory.

In Figure~\ref{fig:growing_lengths}-top, we report test accuracy from a fully backpropagated LSTM baseline, \rlstm{}, and \plstm{} on all levels. Since StanfordDogs is an even more challenging classification problem compared to CIFAR10, pursuing useful accuracy with non-convolutional models is not our main goal. We instead examine the relative robustness of different methods when input sequences get longer. All models are evaluated after 100 epochs of training, with an additional 20 epochs of pretraining for models with auxiliary loss.

Using the unsupervised auxiliary losses, we are able to obtain much better results compared to other methods. Figure~\ref{fig:growing_lengths}-top shows that both \rlstm{} and \plstm{} exhibit the strongest resistance to the growing difficulty, while an LSTM trained with full backpropagation is slow to improve and produces no better than random predictions when the input sequence length reaches the 9\,000 mark. After the 12\,000 mark, memory constraint is exceeded for this model. At the same time, there is virtually no accuracy loss in \rlstm{} going from 12\,000 to 16\,000 element long sequences. 

The gradient truncation in \rlstm{} and \plstm{} also offers a much greater computational advantage as sequence length gets arbitrarily large. Figure~\ref{fig:growing_lengths}-bottom illustrates the time to finish one training step for each recurrent model. LSTM takes 4 seconds at the 1600 mark and quickly stretches out to 26 seconds at the 12\,000 mark. With the same computational resource, our proposed methods stay under 3 seconds and grow up to only around 8 seconds at the end, processing a batch of sequences with lengths more than 16\,000.

\subsection{Comparing with Transformer} 
\label{sec:transformer}

In this set of experiments, we explore how well our proposed recurrent models fare with those that utilize a self-attention mechanism. As noted in the introduction, these models require random access to the entire sequence at inference time, so are very quick to become infeasible as sequences get longer (such as the PTB LM dataset).

\subsubsection{MNIST, pMNIST, and CIFAR10}

On MNIST and pMNIST, Transformer outperforms our best model as shown in Table~\ref{transformer_mnist_cifar}. On CIFAR10, however, Transformer performance drops significantly - worse than most recurrent models on this dataset. 

\begin{table}[htb]
\caption{Test accuracy (\%) on MNIST, pMNIST and CIFAR10.}
\label{transformer_mnist_cifar}
\vskip 0.15in
\begin{center}
\begin{small}
\begin{tabular}{lccc}
\toprule
 & MNIST & pMNIST & CIFAR10 \\
\midrule
\rlstm{} Full BP & 98.4 & 95.2 & 72.2 \\
Transformer & 98.9 & 97.9 & 62.2 \\
\bottomrule

\end{tabular}
\end{small}
\end{center}
\vskip -0.1in
\end{table}

We additionally evaluate results from T-DMCA~\cite{liu18}, though strictly speaking, this is an unfair comparison since the T-DMCA adds convolutions at each self-attention layer. Compared to the Transformer, T-DMCA is more memory efficient as it utilizes local-attention and memory-compressed attention.\footnote{In our experiments, Transformer-DMCA consists of 5 alternating layers of unmasked local attention and memory compressed attention, with all hidden sizes and filter sizes set to 128.} Results indicate that T-DMCA performs better than Transformer on MNIST (99.3\%) and CIFAR10 (73.0\%). On pMNIST where there is no spatial locality to be exploited by convolution, T-DMCA achieves 97.6\% accuracy, slightly worse than that of Transformer.

\subsubsection{StanfordDogs}

Similar to the previous section, we transfer the same hyper-parameter settings of Transformer to much longer sequences, created using StanfordDogs dataset. As shown in Figure~\ref{fig:growing_lengths}, Transformer starts with almost twice the accuracy of \rlstm{} or \plstm{}, but this performance quickly degrades at a much higher rate as input sequences get longer. Specifically, Transformer performs worse than our methods after the 3000 mark and end up only slightly better than random prediction around the 9200 mark. Its training using the same resource also becomes infeasible after this point.

Note that our proposed method is orthogonal to most models that process sequences. Incorporating our technique to any scalable Transformer variant will therefore likely result in significant improvements. Our current work, however, focuses on improving recurrent networks and therefore leaves this option for future exploration.

\subsection{Classifying DBpedia documents at character level}

We explore how well truncated BPTT and the auxiliary losses can do on sequences of discrete data (text), where previous methods already reported remarkable accuracy. For this task, the DBpedia dataset is chosen as it provides a large and carefully curated set of clean Wikipedia texts and no duplication. In our experiments, each document in the dataset is processed at character level~\cite{zhang2015text}. This makes the average sequence length 300, with 99\% of the training examples are under 600 elements long.

\begin{table}[htb]
\caption{Test error rate (\%) on character-by-character DBpedia.}
\label{tab:dbpedia_truncate}
\vskip 0.15in
\begin{center}
\begin{small}

\begin{tabular}{lc}
\toprule
 & Test error \\ 
\midrule
LM-LSTM Truncate100 & 4.04 \\
SA-LSTM Truncate100 & 3.89 \\
\midrule
\rlstm{} 20x15 Truncate100 & 3.84 \\ 
\textbf{\plstm{} Truncate100} & \textbf{2.85} \\
\bottomrule
\end{tabular}
\end{small}
\end{center}
\vskip -0.1in
\end{table}

To explore how well auxiliary losses can help with limited backpropagation, supervised gradients are truncated to only 100 time steps, while anchored subsequences are sampled with length $l=300$. Similar to ~\citet{dai2015semi}, we did not perform joint-training since it slightly degrades performance on this large dataset, all other hyper-parameters are reused. We also test \rlstm{} with the 20-sample setting, a full BPTT trained LSTM baseline and truncated LM-LSTM and SA-LSTM ~\cite{dai2015semi} baselines.

As can be seen in Table~\ref{tab:dbpedia_truncate}, auxiliary losses with truncated BPTT can significantly outperform the LSTM baseline by more than 10\% absolute accuracy. Our methods also have better results than truncated LM-LSTM and SA-LSTM. We conjecture that this comes from the combination of more randomness and truncation in our training process.

\begin{table}[htb]
\caption{Test error rate (\%) on character-by-character DBpedia.}
\label{tab:dbpedia}
\vskip 0.15in
\begin{center}
\begin{small}
\begin{tabular}{lc}
\toprule
 & Test error \\ 
\midrule
LSTM Full Backprop~\cite{dai2015semi}    & 13.64 \\ 
char-CNN~\cite{zhang2015text} & 1.66 \\
CNN+RNN~\cite{xiaocrnn16} & 1.43 \\
29-layer CNN~\cite{ConneauVDCNN16} & 1.29 \\
LM-LSTM~\cite{dai2015semi} & 1.50 \\
SA-LSTM~\cite{dai2015semi} & 2.34 \\
\midrule
\rlstm{} n=20, l=15, 2 layers 512 units & 1.40 \\
\bottomrule
\end{tabular}
\end{small}
\end{center}
\vskip -0.1in
\end{table}

When trained without restriction of model size and gradient truncation, \plstm{} performs on par with other strong baselines that operate on character-level (Table~\ref{tab:dbpedia}). Specifically, \rlstm{} with $n=20$ and $l=15$  significantly outperforms a full auto-encoder in SA-LSTM, ranking only behind Very-deep CNN with 29 layers.

\section{Analysis}

\subsection{Shrinking supervised BPTT length}

Given the clear trend demonstrated in previous sections, it is natural to ask the question of how much longer the input has to grow before \rlstm{} and \plstm{} becomes untrainable. To simulate this effect without growing sequence length indefinitely, we instead keep the input sequence length fixed, while truncating backpropagation incrementally. We perform experiments on CIFAR10 and start shrinking the BPTT length from 300 down to 1 - where gradients from the classification loss have minimal impact on the main LSTM.

\begin{figure}[htb]
\includegraphics[width=0.9\columnwidth, clip=true, trim= 0 5 0 0]{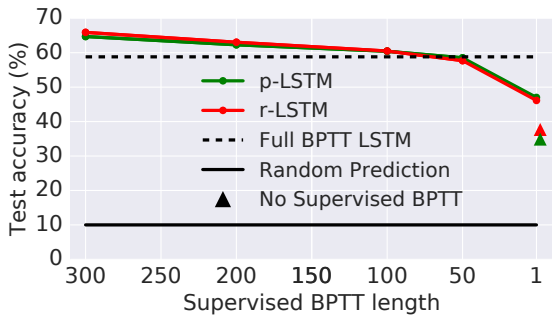}
\caption{Effects of shrinking supervised BPTT length.}
\label{fig:shrinking_bp}
\end{figure}

Results in Figure~\ref{fig:shrinking_bp} shows that \rlstm{} and \plstm{} can afford a reduction of another 200 BPTT steps, while still being able to generalize better than a fully trained LSTM. Moreover, by applying gradients on only 50 steps -- less than 5\% of the total input steps, \rlstm{} and \plstm{}'s accuracy can still approximate their fully trained counterpart. At the extreme point of one-step backpropagation, both \rlstm{} (46.1\%) and \plstm{} (47.0\%) perform commendably well.

Going one step further, we prevent classification gradients from updating the main LSTM, thereby training it completely unsupervised. By doing so, we attempt to address the question of why the human brain can understand long sequences of events, even though BPTT is argued biologically implausible~\cite{Bengio15bioPlausibleML}. Results from Figure~\ref{fig:shrinking_bp} indicate that both \rlstm{} (37.7\%) and \plstm{} (34.9\%) can still classify unseen data with far-from-random accuracy.

\subsection{Multiple reconstructions with fixed BPTT cost} \label{sec:sample_frequency}

So far, we only adopt the simplest setting of $n=1$ reconstruction sample per sequence. One can also tune this hyper-parameter for even better results. We explore this option to improve one and zero step supervised BPTT. 

To keep the total cost of backpropagation approximately the same with previous experiments, we gradually increase $n$ and shrink each subsequence length $l$ proportionately. We also set the unsupervised BPTT truncation to be $l$. In Table~\ref{tab:sample_rate}, we report results obtained with five different sampling frequencies, ranging from 10 to 200 samples.

\begin{table}[htb]
\caption{Classification test accuracy (\%) on CIFAR10 with varying sample frequency and fixed backpropagation cost.}
\label{tab:sample_rate}
\vskip 0.15in
\begin{center}
\begin{small}

\begin{tabular}{cccc}
\toprule
 & & \multicolumn{2}{c}{Supervised BPTT length} \\
\cmidrule(lr){3-4}
$n$ & $l$ & 1 & 0 \\
\midrule
1 & 600 & 46.0 & 37.4 \\
10 & 60     & 46.0 & 40.6  \\
\textbf{20} & \textbf{30}  &  \textbf{48.0} & \textbf{41.6}  \\
50 & 12  &  47.7 & 41.0  \\
100 & 6 & 47.2 & 40.1  \\
200 & 3 &   46.2 & 37.9 \\
\bottomrule
\end{tabular}
\end{small}
\end{center}
\vskip -0.1in
\end{table}

We indeed observed accuracy gain on the test set across almost all frequencies of sampling. Interestingly, there is a peak at 20 samples per sequence and the accuracy gain starts decaying from this point in both directions. In other words, it is harmful to sample too little, or sample too many at the cost of very little backpropagation. Comparing these two extremes, we observe slightly better accuracy with many small reconstructions than one big reconstruction. 

At sampling frequency 20, for single time step backpropagation, we obtain an increase of 2.0\%. For completely unsupervised training (no backpropagation on the main LSTM), there is a remarkable increase of 4.0\%. This increase implies that \rlstm{} has great potential to improve on long sequences, with relatively few supervised gradients, as long as one is able to afford tuning extra hyper-parameters.

We explore this potential on StanfordDogs by retraining \rlstm{} with sampling frequency 20 (\rlstm{} 20$\times$30) on all 8 levels. As shown in Figure~\ref{fig:growing_lengths}-top, this best performing setting found on CIFAR10 generalizes to all difficulty levels of StanfordDogs. Namely, \rlstm{} 20$\times$30 closes the gap with Transformer on shorter sequences, and stays at this top position throughout, outperforming all other recurrent models as well as Transformer by a large margin starting from the 3000 mark .

Furthermore, by independently sampling several segments of equal length, one can batch them to utilize data parallelism and subsequently speed up the training process even more. This is illustrated in Figure~\ref{fig:growing_lengths}-bottom, where single-batch training time of \rlstm{} 20$\times$30 consistently stays lower than that of any other recurrent method.

\subsection{Regularization and Optimization Advantages from Unsupervised Losses} \label{sec:reg_or_opt}

With a significant gap between \rlstm{}/\plstm{} and a fully-trained LSTM on almost all benchmarks, we ask whether it is regularization or optimization advantage that is added by truncated BPTT and our auxiliary losses. At any point during training, we identify optimization advantage when training accuracy with auxiliary losses are much better than that of the baseline, while the corresponding improvement on test set is not as significant. On the other hand, if our models generalize better while being harder or insignificantly easier to train, the improvement comes from regularization.

\begin{figure}[htb]
\includegraphics[width=0.8\columnwidth, clip=true, trim= 0 10 0 0]{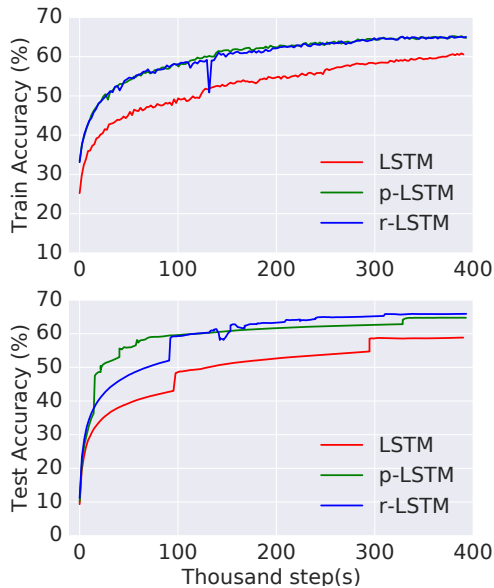}
\caption{Effects of auxiliary losses on training/testing accuracy.}
\label{fig:testing_curves}
\end{figure}

Figure~\ref{fig:testing_curves} shows the training/testing accuracy of \rlstm{}, \plstm{} and an LSTM during training. \rlstm{} and \plstm{} training curves trace each other almost identically throughout, while \rlstm{} gives better result on testing data. This implies that \rlstm{} regularizes better than \plstm{}.

Comparing to the LSTM baseline, \rlstm{} and \plstm{} start off with much higher training accuracy while having the same testing accuracy (10\%). This reveals the significant improvement from unsupervised pretraining for both \rlstm{} and \plstm{}'s optimization. Gradually throughout the training process, this optimization gap with the baseline becomes smaller, while the corresponding difference in test accuracy becomes relatively bigger.

We therefore conclude that both types of pretraining bring optimization advantages at early stages of training. Later on, minimizing the semi-supervised loss creates a regularization effect that quickly takes over until the end.

\subsection{Ablation Study}

\begin{figure}[htb]
\includegraphics[width=\columnwidth, clip=true, trim= 0 10 0 10]{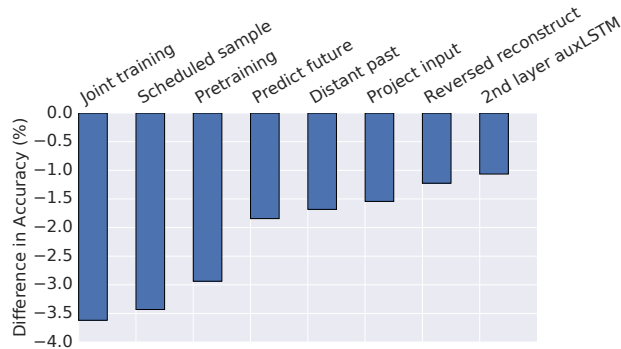}
\caption{Ablation analysis of \rlstm{} performance.}
\label{fig:ablation}
\end{figure}

In this section, we evaluate the relative contribution of different factors to \rlstm{}'s performance. Here we test each factor by turning it off and retraining the model from scratch on CIFAR10, using the same random seed. Firstly, as reported in Table~\ref{mnist_cifar}, eliminating the auxiliary loss and leaving the main LSTM with a truncation of 300 BPTT steps cause a loss of nearly 17\% in test accuracy. With the auxiliary loss in effect, Figure~\ref{fig:ablation} shows the results when turning off other parts from the original full setting.

\textbf{Jointly training unsupervised and supervised loss is most important}, with a corresponding loss of more than 3.6\% reported. As long as joint training is in effect, pretraining is slightly less important than applying Scheduled Sampling for the auxiliary LSTMs.

\textbf{More randomness is better}. Instead of only reconstructing the immediate past, allowing reconstruction segments to be randomly sampled in distant past gives almost a 2\% accuracy gain. Allowing a part of the sampled segment to spread over to the anchor point's future also gives a boost.

Other improvements come from embedding input pixels to the same dimensionality as the LSTM's hidden size, reversing the order of reconstruction and stacking a second layer on the LSTM that receives outputs from the anchor point.

\section{Conclusion}
In this paper, we have presented a simple approach to improve the learning of long-term dependencies in RNNs. An auxiliary loss was added to the main supervised loss to offer two main benefits. 
First, it induces a regularization effect, allowing our models to generalize well to very long sequences, up to length 16\,000.
Second, it provides computational advantages as the input sequence gets very long, so that
one only needs to backpropagate for a small number of time steps to obtain competitive performance. In the extreme cases where there is little to no backpropagation, our models perform far better than random predictions.

On a comprehensive set of benchmarks, ranging from pixel-by-pixel image classification (MNIST, pMNIST, CIFAR10, StanfordDogs) to character-level document classification (DBpedia), our models have demonstrated competitive performance over strong recurrent baselines (iRNN, uRNN, LM-LSTM, SA-LSTM) and non-recurrent ones such as Transformer, CRNN, VDCNN. For long sequences, our results are superior despite using much fewer resources.

We anticipate that this simple technique will be widely applicable to online learning systems or ones that process unusually long sequences.

\bibliography{main}
\bibliographystyle{icml2018}

\end{document}